\documentclass[conference]{IEEEtran}
\IEEEoverridecommandlockouts
\pdfoutput=1
\usepackage{cite}
\usepackage{amsmath,amssymb,amsfonts}
\usepackage{algorithmic}
\usepackage{graphicx}
\usepackage{textcomp}
\usepackage{xcolor}
\usepackage{diagbox}
\usepackage{multirow}
\def\BibTeX{{\rm B\kern-.05em{\sc i\kern-.025em b}\kern-.08em
    T\kern-.1667em\lower.7ex\hbox{E}\kern-.125emX}}
\begin{document}

\title{Spatial Transformer Network with Transfer Learning for Small-scale Fine-grained Skeleton-based Tai Chi Action Recognition
\thanks{This work is supported by the National Natural Science Foundation of China (No. 61876054)}
}

\author{
	\IEEEauthorblockN{Lin Yuan, Zhen He, Qiang Wang, Leiyang Xu, Xiang Ma}
	\IEEEauthorblockA{\textit{Department of Control Science and Engineering}\\
		\textit{Harbin Institute of Technology, Harbin, China} \\
		\textit{\_eunseo\_v@hit.edu.cn, hezhen@hit.edu.cn, wangqiang@hit.edu.cn, xuleiyang@stu.hit.edu.cn, maxianghit@163.com}
	}
}

\maketitle

\begin{abstract}
Human action recognition is a quite hugely investigated area where most remarkable action recognition networks usually use large-scale coarse-grained action datasets of daily human actions as inputs to state the superiority of their networks. We intend to recognize our small-scale fine-grained Tai Chi action dataset using neural networks and propose a transfer-learning method using NTU RGB+D dataset to pre-train our network. More specifically, the proposed method first uses a large-scale NTU RGB+D dataset to pre-train the Transformer-based network for action recognition to extract common features among human motion. Then we freeze the network weights except for the fully connected (FC) layer and take our Tai Chi actions as inputs only to train the initialized FC weights. Experimental results show that our general model pipeline can reach a high accuracy of small-scale fine-grained Tai Chi action recognition with even few inputs and demonstrate that our method achieves the state-of-the-art performance compared with previous Tai Chi action recognition methods.
\end{abstract}

\begin{IEEEkeywords}
Small-scale fine-grained action recognition, Tai Chi action, Transfer learning
\end{IEEEkeywords}

\section{Introduction}
Human action recognition raises a lot of interest among researchers and has been widely applied in scenarios including human-machine interaction and remote monitoring. Skeleton-based human action datasets have been more used as network inputs for their convenience in data collection and low storage consumption. With the development of portable devices and the advancement of transmission efficiency of data, lots of motion-capture systems can easily obtain the 3D skeleton sequences such as Microsoft Kinect\cite{zhang2012microsoft} and Perception Neuron\cite{baumann2017perception}. Microsoft Kinect V2 contains a depth sensor, a color camera, and a four-microphone array that provide full-body 3D motion capture, facial recognition, and voice recognition capabilities and it can easily extract multiple human motions with pose estimation algorithm, but the accuracy is affected by occlusion. Perception Neuron is an inertial sensor-based motion-capture system and each sensor measures its orientation and acceleration using a gyroscope, a magnetometer, and an accelerometer. The motion of the actor can be accurately calculated by wearing the whole system but it can meanwhile only get one actor's motion. 

With the advent of these portable devices, human action recognition using 3D skeleton sequences has attracted more attention in recent years and lots of remarkable approaches have been proposed. In the early years, most studies in skeleton-based action recognition created hand-crafted features to represent the relative 3D rotation and translations among joints\cite{vemulapalli2014human}, but these methods were not general and cannot fully capture all features of human motion. With the development of deep neural networks, deep learning methods have been applied in action recognition. Some works treated 3D skeleton sequences as 2D pseudo-images which represent the skeleton joints and temporal dynamics respectively in rows and columns, and use Convolutional Neural Network-based (CNN-based) methods to extract features from input pseudo-images\cite{ke2017new}\cite{liu2017enhanced}. Skeleton-based human action can also be regarded as a sequence of skeletons and utilize Long Short-Term Memory (LSTM)\cite{hochreiter1997long} to model temporal dependencies among skeleton frames \cite{du2015hierarchical}\cite{song2018spatio}. The above methods ignore the dependencies and break the physical connections among joints, which are not always optimal for all human actions.

\begin{figure*}[t]
	\centering
	\includegraphics[width = 0.9\textwidth]{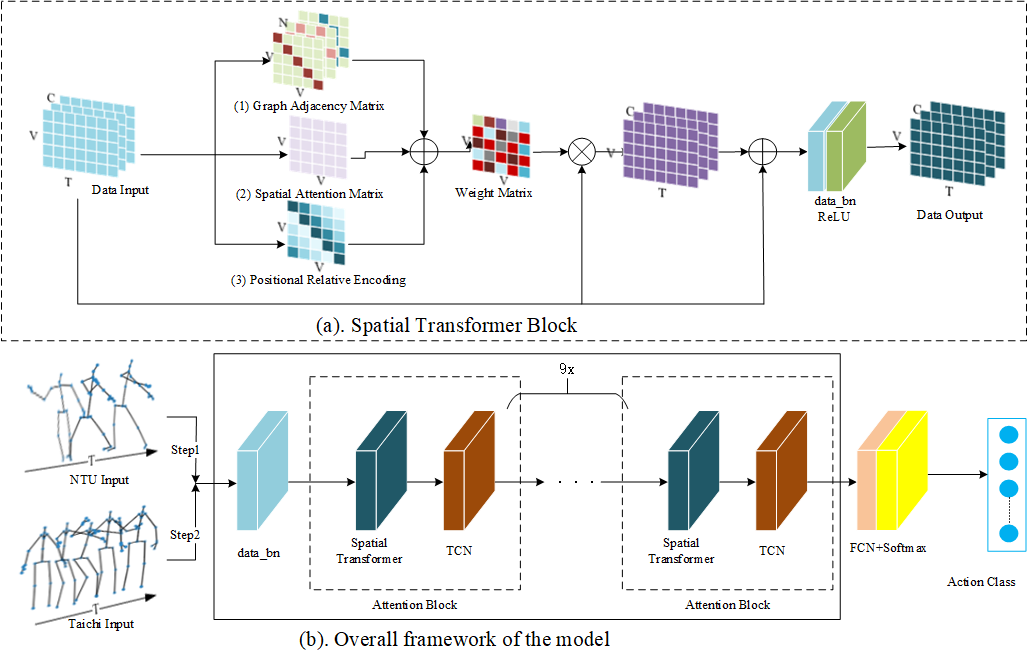}
	\caption{Illustration of proposed transfer learning network. 
		\textbf{(a)} Spatial Transformer Block, 
		\textbf{(b)} Overall framework of the model.}
	\label{network}
\end{figure*}

3D skeleton sequences are essentially graph-structured data. Inspired by CNN, \cite{kipf2016semi} designed an efficient variant of CNN which operate directly on graphs named Graph Convolutional Network (GCN). \cite{yan2018spatial} first leveraged GCN to design a spatial-temporal GCN (ST-GCN) which can automatically capture the patterns embedded in the spatial configuration of the joints as well as their temporal dynamics. \cite{shi2020skeleton} proposed a novel multi-stream attention-enhanced adaptive graph convolutional neural network including fixed structural graph matrix, learnable structural graph matrix, and actional graph matrix varied from input human motion. Bone information and joint information were simultaneously used to form a two-stream framework to improve the recognition accuracy. The above methods have the trend that the weight matrix should be formed according to the input data. While the Transformer architecture\cite{devlin2018bert} with self-attention module\cite{vaswani2017attention} as its mainstream has become the de-facto standard for natural language processing tasks, its structure has been widely applied in various research fields including computer vision\cite{dosovitskiy2020image} and long sequence time-series forecasting\cite{zhou2021informer}. Recently, Transformer algorithms were also applied in action recognition. \cite{plizzari2021skeleton} proposed a novel spatial-temporal network (ST-TR) which models dependence between joints using the Transformer self-attention operator. \cite{zhang2021stst} added a new multi-task self-supervised learning method based on the spatial-temporal network to improve the robustness of the model. Both models achieves remarkable performance on NTU RGB+D datasets\cite{shahroudy2016ntu}\cite{liu2019ntu} compared to the above methods.

Researchers focus on deep and complex neural networks to achieve better results on large-scale coarse-grained action datasets. For examples, training samples vary from $13K$ in UCF-101 dataset to $392K$ in Kinetics-600 datasets. When directly applying these methods on small-scale datasets, these networks usually show severe over-fitting problems. We also notice that prevailing action recognition datasets contain human daily activities. These coarse-grained actions have obvious inter-class distances, which reduce the difficulty of recognition. Recently, some researches focus on fine-grained action recognition\cite{sun2020finegrained} and have established corresponding datasets to tackle application problems on specific scenarios. \cite{nicora2020moca} created the MoCA dataset consisting 20 fine-grained cooking actions from 3 cameras views for each action. \cite{9053928} published a fine-grained basketball action dataset consisting of annotated basketball game videos. These samples could all belong to the class `play basketball' in coarse-grained datasets but are fine-annotated in the basketball dataset. As can be expected, recognizing fine-grained action is more challenging and has more application scenarios.

In this paper, we establish a small-scale fine-grained Tai Chi action dataset and use deep learning method for our Tai Chi action recognition. Researches on Tai Chi action recognition have meanings in action understanding and action evaluation. Tai Chi action can be treated as a specific action class, each Tai Chi action consists of multiple meta actions and is completed by the coordination of the whole body. Compared with human daily actions, attention needs to be paid to the movement characteristics of different human body parts at different stages, so recognizing Tai Chi action is also a challenging task for the similarity between Tai Chi action classes. As Tai Chi has become an Asian Games event, a good recognition network can help abecedarians to evaluate their activities. Some researches have cared about Tai Chi action recognition works, \cite{sun2017taichi} created a fine-grained Tai Chi dataset consisting video clips collected from website with dynamic background. \cite{dong2019tai} \cite{liu2022tai} applied deep learning methods to recognize their own Tai Chi dataset. These datasets are not avalible at website for comparison and methods are not general for fine-grained action recognition.

Our small-scale open-source Tai Chi dataset will lead over-fitting of the network if directly used as network inputs, As Tai Chi actions and those daily actions are all manifestations of human movement, they share similar physical features. Based on the remarkable performance of Transformer and size limitation of Tai Chi dataset, we propose a transfer-learning transformer network for Tai Chi action recognition illustrated in Fig.\ref{network}. We first select 49 single-actor action classes in the NTU RGB+D dataset, and use Spatial Transformer to pre-train the model to capture common human motion features. Then we freeze the weights of the network except initializing the weights of the last FC layer. Finally, we take Tai Chi action as input to train the last FC layer for Tai Chi action recognition. Fig.\ref{network}(b) shows the overall framework of the model while Fig.\ref{network}(a) gives a detailed illustration of a Spatial Transformer block. Our contributions are three-fold: (1) We analyze the application scenarios of Tai Chi action recognition and merge multiple methods including CNN, GCN and, Transformer to extract different features for Tai Chi action recognition. (2) Due to the size limitation of our Tai Chi dataset, we use a similar human action dataset to pre-train the model for human action features extraction. This can reduce the over-fitting phenomenon of the model to improve the recognition accuracy. (3) Compared to the previous recognition method\cite{xu2020using}, the model of this paper is more concise and reaches a higher accuracy even in the little training sample condition.

\section{Tai Chi Dataset}
\label{sec:dataset}

In this section, we first introduce the collection of Tai Chi dataset. Then we explain how we pre-process the Tai Chi sequence to form a suitable input for the model.

\subsection{Sequence Formation}
\begin{figure}[htbp]
	\centerline{\includegraphics[width = 0.4\textwidth]{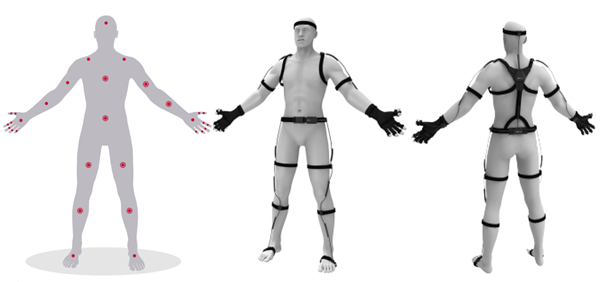}}
	\caption{Neuron position and Sensor wearing combination of Perception Neuron}
	\label{axis neuron}
\end{figure}

For Tai Chi action with a single actor, we use the wearable device Perception Neuron for action data collection. It has up to 32 neurons connected by wires fixed in a specific position, as shown in Fig.\ref{axis neuron}. Each neuron measures its orientation and acceleration, the measured data is sent to the hub which gathers all the data from every connected neuron. The hub then sends that data either over a USB or WiFi connection to a PC to save data or to perform real-time processing. In this experiment, we store the data as a Bounding Volume Hierarchy (BVH) file including the relative rotation of 72 joints. The coordinates of joints in the local coordinate system can be calculated through this file. 

BVH file consists of two parts. The first part defines each joint's name, several channels, and the relative positions between the child joint and its parent joint. This gives a sketch of input sequence. The second part gives the channel data of each joint at every frame including the rotation and translation information, so the coordinates of a skeleton joint in the local coordinate system can be calculated in a chain multiplication with a combined rotation matrix until finding its parent joint same as the root joint. The combined rotation matrix $R$ is calculated by multiplying the matrix of each axis, $R_x$, $R_y$, $R_z$ respectively. Supposing we get the rotation angle of three-axis $\alpha$, $\beta$, $\gamma$ and translation position $T_x$, $T_y$, $T_z$ of a child joint to its parent joint, we get the rotation matrix of each axis as shown in \eqref{RoMat}. 

To calculate the coordinates of a joint, take the left foot joint as an example, assuming the relative position to its parent joint is $x_0, y_0, z_0$, the coordinates in the local coordinate system defined as $x_1, y_1, z_1$ can be calculated in \eqref{coordinate}, where $M$ is the transformation Matrix shown in \eqref{RoMatSum}. A sequence of Tai Chi action ``Preparation'' after the calculation has shown in lower-left corner of Fig.\ref{network}.

\begin{align}
\nonumber R_{x}=\begin{pmatrix}
1 & 0 & 0 \\
0 & cos\alpha & -sin\alpha \\
0 & sin\alpha & cos\alpha
\end{pmatrix} \\
\nonumber R_{z}=\begin{pmatrix}
cos\theta & -sin\theta & 0 \\
sin\theta & cos\theta & 0 \\
0 & 0 & 1
\end{pmatrix} \\
R_{y}=\begin{pmatrix}
cos\varphi & 0 & sin\varphi \\
0 & 1 & 0 \\
-sin\varphi & 0 & cos\varphi
\end{pmatrix}
\label{RoMat}
\end{align}

\begin{equation}
\begin{pmatrix}
x_1 \\
y_1 \\
z_1 \\
1
\end{pmatrix} = M_{Hips}M_{LeftUpLeg}M_{LeftLowLeg}M_{LeftFoot}\begin{pmatrix}
x_0 \\
y_0 \\
z_0 \\
1
\end{pmatrix}
\label{coordinate}
\end{equation}

\begin{equation}
M=\begin{pmatrix}  
R & \begin{matrix}  
T_x \\ 
T_y \\
T_z
\end{matrix}  \\  
\begin{matrix}  
0 & 0 & 0  
\end{matrix}  & 1  
\end{pmatrix}
\label{RoMatSum} 
\end{equation}

\subsection{Sequence Pre-processing}
\begin{figure}[htbp]
	\centerline{\includegraphics[width = 0.4\textwidth]{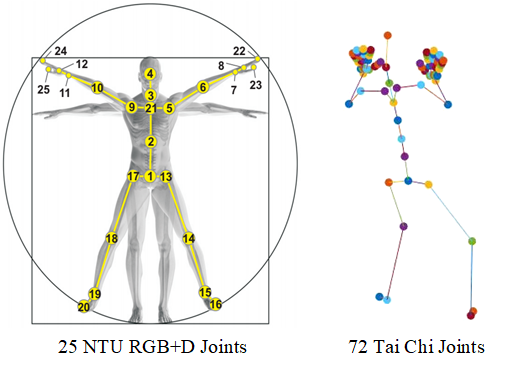}}
	\caption{Joints of NTU RGB+D and Tai Chi dataset}
	\label{joint comparision}
\end{figure}

Our Tai Chi action dataset consists of 10 Tai Chi actions and each action has 20 samples. As for our small-scale Tai Chi dataset, we intend to use the NTU RGB+D dataset to pre-train the model. 25 joints\cite{shahroudy2016ntu} of NTU RGB+D dataset are provided as we get 72 joints of Tai Chi dataset. As illustrated in Fig.\ref{joint comparision}, Tai Chi dataset has abundant hand information. So the process of joint conversion will lose some action information about hands. Then we augment the size of the training Tai Chi dataset to avoid the overfitting of the network. We also normalize the sequence to a fixed angle to reduce the impact of various viewpoints. Finally, we set all sequences to 64 frames by bilinear interpolation.

\section{Transfer Learning Network}
\label{sec:network}

\subsection{Graph Convolutional Network}
An action data can be represented by 2D or 3D coordinates of each human joint in each frame. Our input $X\in \mathbb{R}^{C \times T \times V}$ can be abstracted as $V$ joints representing each skeleton and $T$ frames composing the sequence. As mentioned in model ST-GCN \cite{yan2018spatial}, we bring in the graph adjacency matrix of the skeleton shown in Fig.\ref{network}(a)(1) to capture the structural links between joints. After normalization, we finally get spatial adjacency matrix $A\in\mathbb{R}^{K \times V \times V}$, where $K=3$. In our model, we set matrix $A$ trainable to learn the dynamic structural links between joints, making the model adaptive.

\subsection{Self-attention Mechanism}
The self-attention mechanism can automatically learn the relationships among the input data with variable weight matrix. The function self-attention can be described as mapping a query and a set of key-value pairs to an output, where the query, keys, values, and the output are all vectors. The output is computed as a weighted sum of the values, where the weight assigned to each value is computed by a combination of the query and the corresponding key. Similar to \cite{plizzari2021skeleton}, we bring in the spatial self-attention to our model to extract features embedding the relations between joints. Given the input sequence $X\in\mathbb{R}^{C\times T\times V}$, we take $v_{ti}$ at time $t$ with joint $i$ as example. We first compute a query vector $q_i^t\in\mathbb{R}^{d_q}$, a key vector $k_i^t\in\mathbb{R}^{d_k}$ and a value vector $v_i^t\in\mathbb{R}^{d_v}$ using input vector $v_{ti}$ multiplied by trainable parameters $W_q\in\mathbb{R}^{C_{in}\times d_q}$, $W_k\in\mathbb{R}^{C_{in}\times d_k}$, $W_v\in\mathbb{R}^{C_{in}\times d_v}$ which shared across all nodes. So, for any pair $\{v_{ti},v_{tj}\}$, we get a query-value score $a_{ij}=q_i^t \cdot (k_j^t)^T/\sqrt{d_k}$. For joint $i$, it can get a query-value score vector including all joints. The attention score vector for joint $i$ is shown in \eqref{attention score vector} after the softmax function.
\begin{equation}
B_{i, :} = softmax([a_{i1}, a_{i2}, \dots, a_{iV}]) \in \mathbb{R}^{1\times V}
\label{attention score vector}
\end{equation}

Concatenating each joint's attention score vector, we form Spatial Attention Matrix $B\in\mathbb{R}^{V\times V}$ in Fig.\ref{network}(a)(2). In our model, multi-head attention is applied by splitting the input channel to $N_h$ folds, the self-attention outputs of each fold are concatenated together and pass through an FC layer to form the final output features.

\subsection{Spatial Transformer Block}
The structure of our spatial transformer block is shown in Fig.\ref{network}(a). Besides graph adjacency matrix and spatial attention matrix, positional relative encoding matrix shown in Fig.\ref{network}(a)(3) is added to encode the different joint information, as it is obvious that joints of hand and foot have different meanings even they are in the same location. In the Spatial Transformer block, input data are multiplied by the weight matrix to extract spatial features of the skeleton. Residual connections are added to stabilize the model. Batch normalization and a ReLU\cite{agarap2018deep} layer are applied to extracted features to form the spatial transformer block output.

\subsection{Framework of Model Pipeline}
Illustrated in Fig.\ref{network}(b), as for a batch of skeleton sequences $X\in\mathbb{R}^{N\times C\times T\times V}$, we first use a two-dimensional batch normalization to reduce internal covariate shift. Then a hierarchy of attention blocks is applied to extract spatial and temporal features of the normalized sequence. Each block contains Spatial Transformer and Temporal Convolution Network(TCN). We set a 2D convolutional layer as the TCN while Spatial Transformer has mentioned before. We reshape the input $X$ to $X_s\in\mathbb{R}^{NT\times V \times C_s}$ to extract features of each frame. In TCN, a 2D kernel $(t,1)$ is applied on the input $X_t\in\mathbb{R}^{N\times C_t\times T\times V}$ for temporal modeling. Extracted features finally pass through the FC layer and softmax function to predict the probability of each class. 

\section{Experimental Details}
In this section, we first give a brief introduction of two datasets. Then we state the experimental settings and experiment contents. Finally, we give the recognition results of the experiment and the comparison to conventional methods.

\subsection{Experiment Datasets}
NTU RGB+D dataset is used to pre-train the network. The NTU RGB+D dataset \cite{shahroudy2016ntu} is a large-scale dataset for 3D human action collected by Microsoft Kinect V2. It consists of 60 different action classes including single-player actions and multi-player interaction. In this work, we select the first 49 classes which are all single-player actions and use Cross-View evaluation (X-View) to pre-train the model. The denoised dataset has 30349 training and 15115 testing samples in X-View, split according to the different camera views.

\begin{figure}[htbp]
	\centerline{\includegraphics[width = 0.5\textwidth]{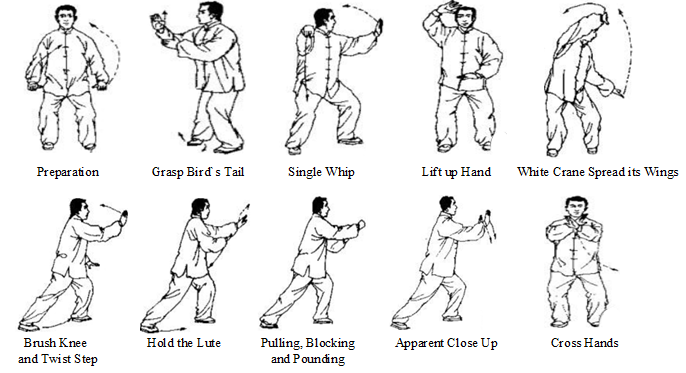}}
	\caption{Illustration of 10 Tai Chi actions}
	\label{wutaichi}
\end{figure}

Our small-scale fine-grained Tai Chi dataset has 10 Tai Chi actions with 20 samples in each action, and our dataset is available at https://cloud.hit605.org/s/taichi. Representations of each action's iconic frame is illustrated in Fig.\ref{wutaichi}. As shown in Fig.\ref{wutaichi}, fine-grained Tai Chi actions are composed of a series of meta action and are hard to recognize with one frame. For example, `Brush Knee and Twist Step', `Hold the Lute', `Pulling, Blocking and Pounding', `Apparent Close Up' share common human poses with little different hand gestures. Limitation of the small-scale dataset and close inter-class distances increases the recognition difficulties.

\subsection{Experiment Settings}
We split 10\%, 30\%, 50\%, 70\% of Tai Chi action samples as training samples respectively for action recognition. We also augment the training dataset up to 8 times to enrich the dataset samples and reduce over-fitting.

Our model is composed of 9-layer attention blocks and the channel dimension of each block is 64, 64, 64, 128, 128, 128, 256, 256, 256. In Spatial Transformer block, the number of multi-head attention is 8 and the dimension of $d_q, d_k, d_v$ are all $0.25\times C_{out}$. Using Pytorch \cite{paszke2019pytorch} framework, we first pre-train our model using NTU RGB+D dataset for a total of 100 epochs with batch size 32 and SGD as optimizer. The initial learning rate is 0.1 and is reduced by a factor of 10 at epoch 50, 70 and 90. We also use distributed data-parallel framework (DDP) to accelerate the training of the model as 8 GPUs are simultaneously used for calculation.

While finishing the training of the NTU RGB+D dataset, we freeze the weights of the network except for the FC layer and initialize the FC weights, action classes reduce from 49 to 10. All configuration parameters are the same as NTU dataset training. 12 different training dataset settings are used to demonstrate the effectiveness of the pre-trained model.

\subsection{Experiment Results and Analysis}

The recognition accuracy of these settings in shown in Table.\ref{taichi acc}. As star bold labeled in Table.\ref{taichi acc}, an action sample `Pulling, Blocking and Pounding' is misclassified into action `Apparent Close Up'. As shown in Fig.\ref{wutaichi}, both action share similar body trend and have same motion direction during the action. Owing to our pre-trained process, we have to discard some hand information in Tai Chi action for skeleton joint pairing, which may lead to misclassifying.

\begin{table}[htbp]
	\caption{Tai Chi Action Recognition Accuracy}
	\renewcommand\arraystretch{1.2}
	\setlength{\tabcolsep}{5MM}{
		\begin{center}
			\begin{tabular}{|c|c|c|c|}
				\hline
				\diagbox{Ratio}{Acc}{Aug} & 2 &4 &8 \\ \hline
				10\% & 87.22\% & 87.22\% & 85.56\% \\ \hline
				30\% & 89.29\% & 90.71\% & 90.71\% \\ \hline
				50\% & 92.00\% & 93.00\% & 96.00\% \\ \hline
				70\% & 95.00\% & \textbf{98.33\%*} & 96.67\% \\ \hline
			\end{tabular}
			\label{taichi acc}
	\end{center}}
\end{table}

We first analyze the effect of sample augmentation. The performances of recognition increase notably when the augmentation factor increases to 4 but may drop a little when continuously raise to 8, especially for the small amount of training samples. This happens because too many similar samples restrict the universality of the model and make the model overfit.

From the perspective of training ratio, we find that even with 10\% of training samples, that is only 2 samples included during training, the model has a remarkable recognition effect. With the growing of training ratio, the accuracy increases steadily, which states the effectiveness of our transfer learning model.

Compared to the conventional method\cite{xu2020using} whose recognition accuracy is shown in Table.\ref{acc convention}, they applied a conventional method SVM to recognize the segmented trajectory as opposed to deep learning methods and didn't perform well under small-scale settings. Our model outperforms their method at all configurations. In our transfer learning model, we don't need to create handcrafted features or segment trajectories, using the sample itself to extract features for action recognition instead and reach higher recognition accuracy.

\begin{table}[htbp]
	\caption{Tai Chi Action Recognition Accuracy Using Conventional Method}
	\renewcommand\arraystretch{1.2}
	\setlength{\tabcolsep}{7MM}{
		\begin{center}
			\begin{tabular}{|c|c|c|}
				\hline
				& \begin{tabular}[c]{@{}c@{}}Training\\ Sample\\ Ratio\end{tabular} & Acc     \\ \hline
				\multirow{3}{*}{\begin{tabular}[c]{@{}c@{}}Single\\ node\end{tabular}} & 25\%                                                              & 80.05\% \\ \cline{2-3} 
				& 50\%                                                              & 90.79\% \\ \cline{2-3} 
				& 75\%                                                              & 93.86\% \\ \hline
				\begin{tabular}[c]{@{}c@{}}Multiple\\ nodes\end{tabular}               & Temples                                                           & 90.45\% \\ \hline
			\end{tabular}
	\end{center}}
	\label{acc convention}
\end{table}

\section{Conclusion and future work}
In this paper, we build a fine-grained Tai Chi dataset with professional Tai Chi performers using a wearable motion capture system. Each Tai Chi action is composed of multiple meta actions with the coordination of whole body parts. Actions among classes have little differences and they are suitable for the fine-grained action recognition task. Compared with creating hand-crafted features for Tai Chi action or other fine-grained action recognition methods, we propose a general model pipeline that can extract discriminative motion features even on small-scale fine-grained datasets, and this method has not yet applied in other fine-grained action recognition research. Experiments demonstrate the our model pipeline can reach steady and accurate performances under a small-scale training dataset. Our future work focuses on improving the recognition accuracy under the condition of a small training set ratio with the combination of Spatial and Temporal Transformer.

%\begin{table}[htbp]
%\caption{Table Type Styles}
%\begin{center}
%\begin{tabular}{|c|c|c|c|}
%\hline
%\textbf{Table}&\multicolumn{3}{|c|}{\textbf{Table Column Head}} \\
%\cline{2-4} 
%\textbf{Head} & \textbf{\textit{Table column subhead}}& \textbf{\textit{Subhead}}& %\textbf{\textit{Subhead}} \\
%\hline
%copy& More table copy$^{\mathrm{a}}$& &  \\
%\hline
%\multicolumn{4}{l}{$^{\mathrm{a}}$Sample of a Table footnote.}
%\end{tabular}
%\label{tab1}
%\end{center}
%\end{table}

\vfill\pagebreak

\bibliographystyle{IEEEtran}
\bibliography{reference}

\end{document}